\documentclass[11pt]{article}

\usepackage[]{acl}
\usepackage{xcolor}
\usepackage{times}
\usepackage{latexsym}
\usepackage{amsmath}
\usepackage{booktabs}
\usepackage{soul}
\usepackage{scalefnt}
\usepackage{amsfonts}
\usepackage{multirow} 
\usepackage{booktabs}
\usepackage{geometry}
\usepackage{array}
\usepackage{tabularx}
\usepackage[most]{tcolorbox}
\usepackage{enumitem}
\usepackage[utf8]{inputenc}

\newenvironment{ttcitemize}{
  \begin{itemize}[leftmargin=*, topsep=2pt, itemsep=1pt, parsep=0pt, partopsep=0pt]
}{
  \end{itemize}
}

\tcbset{
  ttcbox/.style={
    boxsep=4pt,
    left=6pt,
    right=6pt,
    top=6pt,
    bottom=6pt,
    before skip=6pt,
    after skip=6pt
  }
}

\usepackage[T1]{fontenc}
\usepackage[utf8]{inputenc}

\usepackage{microtype}

\usepackage{inconsolata}

\usepackage{graphicx}

\title{Self-Explaining Hate Speech Detection with Moral Rationales}

\author{Francielle Vargas \\ University of Chile \\
\And
Jackson Trager \\  University of Southern California \\
\And
Diego Alves  \\ Saarland University \\
\AND
Matteo Guida \\ University of Melbourne \\
\And
Surendrabikram Thapa \\ Virginia Tech \\
\And 
Berk Atil \\ Pennsylvania State University \\
\AND
Daryna Dementieva \\ Technical University of Munich \\ % \\ Munich Center for Machine Learning \\
\And
Andrew Smart \\ Google Research\\
\And
Ameeta Agrawal \\ Portland State University \\
}
\begin{document}
\maketitle
\begin{abstract}
Existing hate speech detection models are often opaque and rely on surface-level lexical cues, which makes them vulnerable to spurious correlations and limits robustness, interpretability and cultural contextualization. We propose Supervised Moral Rationale Attention (SMRA)\footnote{Dataset, annotator disagreements, and code are available at \url{https://github.com/franciellevargas/SMRA}}, the first self-explaining hate speech detection framework to incorporate moral rationales as direct supervision for attention alignment. Based on Moral Foundations Theory, SMRA aligns token-level attention with expert-annotated moral rationales, guiding models to attend to morally salient spans. Unlike prior rationale-supervised or post-hoc approaches, SMRA integrates moral rationale supervision directly into the training objective, producing inherently interpretable and contextualized explanations. To support our framework, we also introduce HateBRMoralXplain, a Brazilian Portuguese benchmark dataset annotated with hate labels, moral categories, token-level moral rationales, and socio-political metadata. Across binary hate speech detection and multi-label moral sentiment classification, SMRA consistently improves performance while enhancing both faithful and plausible explanations. Although explanations become more concise, sufficiency decreases, indicating more compact and informative rationales. Fairness remains stable, suggesting that improvements in explanation quality do not introduce significant bias trade-offs. \footnote{\textbf{Warning}: This document contains offensive content.}
 %Overall, SMRA strengthens both classification quality and explainability compared to models without supervised attention.
%substantially improves explanation faithfulness and interpretability (Token F1: +33.2\%; Comprehensiveness: +7.2\%), while preserving predictive performance (Macro F1: +2.0\%; AUROC: +0.2\%) and maintaining stable group-based fairness (GMB-Sub: +0.8\%). These results demonstrate that supervised moral attention enables faithful, transparent, and culturally grounded hate speech detection while maintaining effectiveness and fairness.
\end{abstract}

%-----------------------------------------
\section{Introduction}
\label{sec:introduction}
%-----------------------------------------
Despite significant advances in automatic Hate Speech (HS) detection, current approaches remain fundamentally limited. Existing models often encode biases originating from training data \cite{davidson-etal-2019-racial,wiegand-etal-2019-detection}, stereotypical associations learned by hate speech classifiers \cite{davani2023,vargas-etal-2023-socially}, and annotation processes \cite{sap-etal-2019-risk}, which are frequently shaped by annotators’ subjective judgments and sociocultural backgrounds \cite{ mostafazadeh-davani-etal-2024-d3code,tonneau-etal-2024-languages,abercrombie-etal-2023-temporal,sap-etal-2022-annotators,vargas-etal-2022-hatebr,Polettoetall2021}. Such biases can lead to systematic disparities in model behavior, resulting in the unfair treatment or over-targeting of marginalized groups and producing negative social consequences when deployed at scale \cite{blodgett-etal-2020-language,wiegand-etal-2019-detection,sap-etal-2019-risk,davidson-etal-2019-racial}.

\begin{figure}[!t]
\centering
\includegraphics[width=0.40\textwidth]{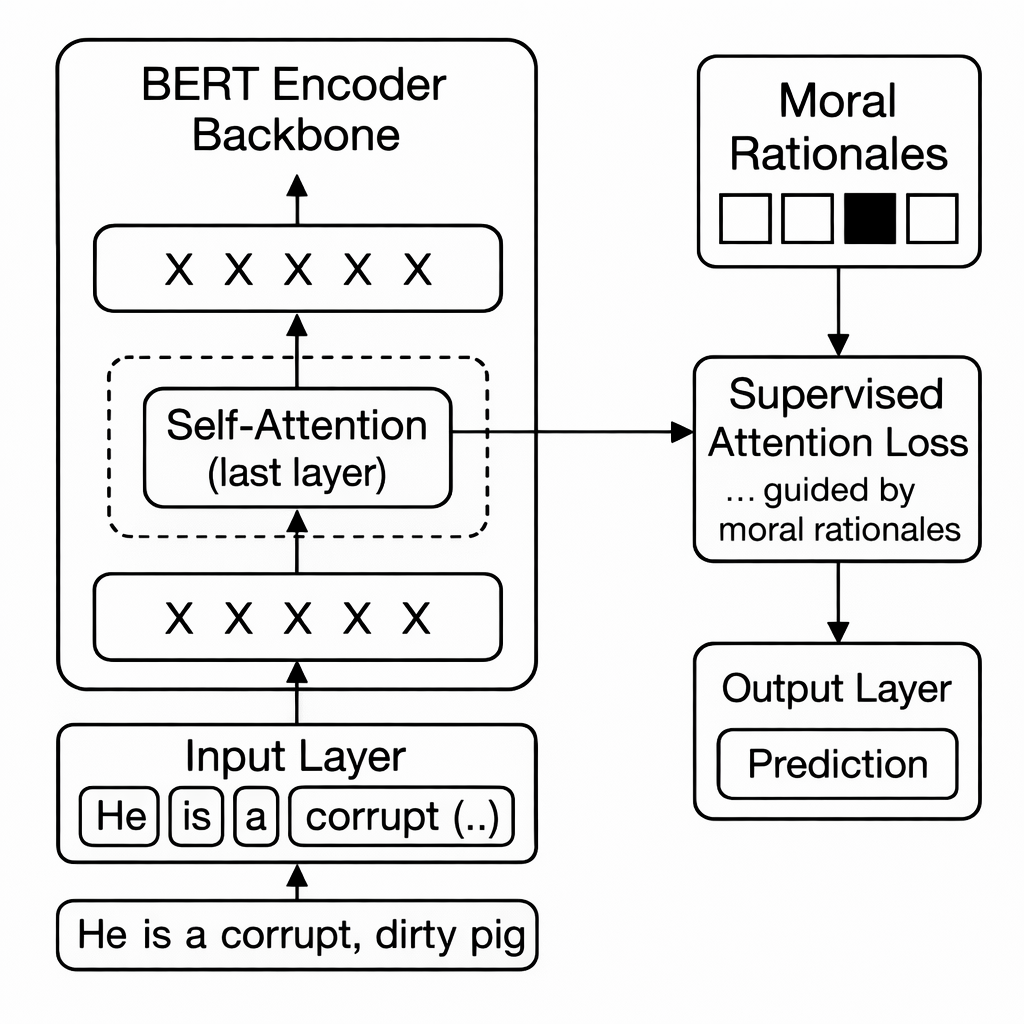}
\caption{Supervised Moral Rationale Attention (SMRA) for Self-Explaining Hate Speech Detection.}
\label{fig:smra}
\end{figure}

Given that these models may inherit biases from their training data, explainable hate speech detection models \cite{salles-etal-2025-hatebrxplain,eilertsen2025aligningattentionhumanrationales,mathew2021hatexplain} are crucial for ensuring transparency, diagnosing biases, and enabling trustworthy detection models across diverse languages and cultural contexts. Existing explainable HS detection approaches can be broadly categorized into two interpretability paradigms: post-hoc and self-explaining models. Post-hoc approaches generate explanations after the prediction, relying on external attribution or surrogate techniques to approximate the model’s reasoning \cite{trager2025mftcxplain,salles-etal-2025-hatebrxplain,cercas-curry-etal-2023-milanlp,mathew2021hatexplain,kennedy-etal-2020-contextualizing}. Moreover, several post-hoc explainability methods remain unfaithful and computationally expensive. In contrast, self-explaining models integrate explanation generation directly into the learning process through inherently interpretable mechanisms, e.g., supervision that aligns model attention with human-annotated rationales, making explanations an integral part of the prediction pipeline \cite{eilertsen2025aligningattentionhumanrationales,kim-etal-2022-hate,mathew2021hatexplain}. 

Furthermore, most existing hate speech detection approaches continue to rely primarily on surface-level lexical features, which are insufficient due to elevated false positive rates \citep{Polettoetall2021, davidson-etal-2019-racial} and their limited ability to account for cultural and contextual factors \cite{lee-etal-2023-hate, zhou-etal-2023-cultural, kim-etal-2024-click, Vargas_Carvalho_Pardo_Benevenuto_2024,wong-2024-sociocultural}. Even when contextual information is available \cite{vargas-etal-2021-contextual,kennedy-etal-2020-contextualizing}, judgments of offensiveness remain deeply culture-dependent and normatively grounded, challenging models that assume stable or universal interpretations of hate speech \cite{tonneau-etal-2024-languages, lee-etal-2023-hate}. More fundamentally, existing models fail to capture normative and moral reasoning, limiting their ability to distinguish culturally contingent hate speech from harmless language \cite{trager2025mftcxplain, tonneau-etal-2024-languages, mostafazadeh-davani-etal-2024-d3code}.

%However, most existing approaches remain focused on surface-level lexical cues, which offensiviness is deeply culutral-context depedent resulting in high rates of false positives \citep{davidson-etal-2019-racial}, belised rarely capture the underlying normative reasoning that drives hateful judgments \cite{tonneau-etal-2024-languages,mostafazadeh-davani-etal-2024-d3code}.

Recent studies provide empirical evidence that moral values constitute a transferable latent representation for hate speech across languages and cultures \cite{trager2025mftcxplain, mostafazadeh-davani-etal-2024-d3code,solovev2023moralized, kennedy2023moral,10.1145/3465336.3475112}.  Cross-lingual studies based on Moral Foundations Theory show that, despite variation in the lexical realization of hateful content across languages, hate speech targets a recurring set of moral violations and rarely occurs in the absence of moral sentiments \cite{trager2025mftcxplain, upadhyaya-etal-2023-toxicity,10.1093/pnasnexus/pgac281}. A recent study by \citet{mostafazadeh-davani-etal-2024-d3code} further shows that disagreements in hate speech annotation are shaped by annotators’ moral values rather than annotation noise. These findings suggest that standard label aggregation and opaque modeling approaches risk encoding dominant moral norms while obscuring minority perspectives. %In contrast, our approach anchors model predictions in moral rationales, rendering value-laden decisions transparent and auditable rather than implicit.

To address these limitations, which risk amplifying existing social and cultural biases, \textbf{we introduce Supervised Moral Rationale Attention (SMRA), the first framework to align model attention with human-annotated moral rationales for self-explaining hate speech detection}. As illustrated in Figure~\ref{fig:smra}, SMRA introduces a normative 
tive regularization into the learning process through a supervised attention loss guided by moral rationales. While lexical markers of hate speech are highly language- and culture-dependent \citep{tonneau-etal-2024-languages}, moral categories and their rationales encode more stable normative structures \citep{atari2023morality, kennedy2021moral}, leading to reduced reliance on spurious correlations and improving robustness, interpretability, and cross-cultural generalization.

Experimental results show that SMRA consistently improves both predictive performance and explanation faithfulness and plausibility across binary hate speech detection and multi-label moral sentiment classification. In binary classification, SMRA improves Accuracy ($+0.0004$) and Macro F1 ($+0.0082$), while substantially enhancing rationale alignment, with gains in IoU F1 ($+0.0743$) and Token F1 ($+0.0503$). In addition, SMRA produces more concise explanations, with lower Comprehensiveness ($-0.0203$) and improved Sufficiency ($-0.0231$), indicating more compact yet still faithful rationales. In multi-label moral sentiment classification, SMRA further improves Macro F1 ($+0.015$), AUROC ($+0.002$), and Token F1 ($+0.241$), demonstrating stronger rationale coverage and interpretability without degrading predictive performance. Fairness remains stable, suggesting that improvements in explanation quality do not introduce significant bias trade-offs.

%We further conduct a large-scale interdisciplinary analysis at the intersection of computational linguistics, moral psychology, and social science, using metadata from Instagram with expert-labeled moral rationales (e.g., gender and political affiliation). 

In addition, we evaluate the effectiveness of large language models (LLMs) for hate speech and moral sentiment classification. Our findings show that incorporating moral rationales leads to notable improvements in hate speech classification performance ($\approx 2$--$3\%$), although LLMs perform poorly on moral sentiment classification, with a maximum F1 of 0.38. Finally, we also introduce HateBRMoralXplain, the third version of the HateBR benchmark dataset for low-resource Brazilian Portuguese \cite{vargas-etal-2022-hatebr,salles-etal-2025-hatebrxplain}. \textbf{To the best of our knowledge, HateBRMoralXplain is the first large-scale expert-annotated corpus of its kind that enriches hate speech data with both moral categories and human-annotated moral rationales}. %It extends HateBR \citep{vargas2022hatebr, Vargas_Carvalho_Pardo_Benevenuto_2024} and HateBRXplain \citep{salles2025hatebrxplain} by incorporating moral categories and moral rationales. The corpus comprises 7,000 Instagram comments from Brazilian political accounts, enabling systematic evaluation of explainable and  morality-aware hate speech detection.

%\subsection{Limitations of Prior Work}
%\mee{this section can be removed; much of this is clear from the Intro}Despite advances in explainable hate speech detection, prior work remains methodologically limited. Post-hoc explanation methods are weakly coupled to model decisions, often unfaithful, and computationally costly, providing little control over the reasoning process. Existing self-explaining approaches partially address these issues but largely rely on lexical rationales, which capture surface patterns of offensiveness rather than the underlying moral and normative structure of hate speech. As a result, such models remain vulnerable to spurious lexical correlations, annotation artifacts, and cultural bias, leading to poor robustness and limited cross-cultural generalization. More fundamentally, treating hate speech as a purely lexical phenomenon obscures its inherently moral and context-dependent nature, constraining both interpretability and real-world applicability.

%\subsection{Contributions}
Our contributions can be summarized as follows:
\begin{enumerate}
    \item We introduce Supervised Moral Rationale Attention (SMRA), the first self-explaining framework that aligns model attention with expert-annotated moral rationales to improve interpretability, robustness, and contextualization in hate speech detection; 
    \item We release HateBRMoralXplain, an extended version of a widely used Brazilian Portuguese hate speech  benchmark, augmenting prior datasets with moral categories, moral rationales, and socio-political metadata; the dataset, models, inter-annotator disagreements, and code are publicly available to facilitate future research; 
    \item We evaluate multiple LLMs on HateBRMoralXplain corpus for hate speech and moral sentiment classification using prompts with different informational components, demonstrating that the inclusion of moral rationales provide statistically significant improvements in hate speech detection performance.
\end{enumerate} %(i) (ii) (iii) 

%------------------------------------------
\section{Related Work}
\label{sec:relatedwork}
%------------------------------------------

%------------------------------------------------
\paragraph{Self-Explaining Hate Speech Detection.}
%------------------------------------------------
 %FERRET \cite{attanasio-etal-2023-ferret} provides a unified framework for generating and evaluating post-hoc explanations for transformer-based models, but such explanations remain weakly coupled to model decisions and often lack faithfulness. 
%A lexicon-based approaches that aggregate VAD features using NRC-VAD and Hurtlex \cite{sariyanto-etal-2025-towards} achieve competitive performance across multiple benchmarks, their explanations stem from deterministic lexical computations rather than learned reasoning. 
Moving toward intrinsic explainability, Supervised Rational Attention (SRA) \cite{eilertsen2025aligningattentionhumanrationales} aligns transformer attention with human rationales and substantially improves rationale faithfulness; however, its reliance on lexical rationales limits robustness, cross-cultural generalization, and the ability to capture deeper moral reasoning. Together, these limitations motivate self-explaining frameworks that explicitly incorporate moral context beyond surface-level cues. \citet{nirmal-etal-2024-towards} present SHIELD for interpretable hate speech detection by incorporating GPT-3.5–generated textual rationales as additional input. The text and rationales are encoded separately using HateBERT and a frozen BERT, respectively, and their representations are concatenated for classification. This design promotes interpretability by grounding predictions in explicit rationale representations. \citet{calabrese-etal-2022-explainable} formulate Intent Classification and Slot Filling (ICSF) task. The approach uses a two-stage BART-based sequence-to-sequence model that first produces a semantic sketch specifying relevant slots (e.g., Target, ProtectedCharacteristic), and then populates these slots with the corresponding text spans. The abuse intent (e.g., Dehumanization, Derogation) is deterministically derived from the filled slots, making the prediction explicitly grounded in extracted evidence. This structured output provides an intrinsic, human-interpretable explanation, rather than a post-hoc rationale. \citet{kim-etal-2022-hate} introduces an intermediate token-level task in which portions of the rationale embeddings are masked, and the model is trained to reconstruct the masked rationale labels using contextual information. This rationale prediction objective, optimized via cross-entropy loss on masked tokens, encourages context-aware reasoning before fine-tuning the model for hate speech classification.

%that incorporate moral rationales for more robust and morally aligned detection.

%[HELPFUL TO ADD LIMITATIONS OF THESE WORKS; CAN ALSO SPLIT THESE INTO METHODS VS. DATASETS AND MENTION THEIR LIMITATIONS]

%------------------------------------------------
\paragraph{Hate Speech with Moral Rationales.}
%------------------------------------------------
Recent work underscores the role of moral rationales in advancing hate speech detection beyond surface-level cues. MFTCXplain \cite{trager2025mftcxplain} introduces a multilingual benchmark with expert annotations of hate labels, Moral Foundations Theory (MFT) categories, and span-level moral rationales, showing that while state-of-the-art LLMs perform reasonably in classification, they struggle to predict moral categories and generate faithful explanations. Complementarily, \citet{mostafazadeh-davani-etal-2024-d3code} demonstrate that although the lexical realization of moral values varies across languages, moral categories provide consistent predictive signals that improve cross-lingual robustness and transfer. Similarly, TWISTED \cite{upadhyaya-etal-2023-toxicity} jointly models toxicity, moral values, and speech acts, thus producing consistent gains across benchmarks. Together, these works indicate that explicitly modeling moral dimensions supports more robust and generalizable hate speech detection while reducing reliance on spurious lexical correlations. Further literature on the relationship between hate and morality is discussed in Appendix~\ref{app:hate_and_moral}.

%------------------------------------------------
\section{HateBRMoralXplain Corpus}
\label{sec:dataset}
%------------------------------------------------
%Following prior work, we argue that hate speech constitutes an extreme form of moralized discourse, in which social groups are evaluated through moral lenses \cite{trager2025mftcxplain,kobbe-etal-2020-exploring}. 
We introduce HateBRMoralXplain, the third version of the HateBR corpus  \citep{vargas-etal-2022-hatebr,Vargas_Carvalho_Pardo_Benevenuto_2024}, extending both the original HateBR dataset and its explainable successor, HateBRXplain \citep{salles-etal-2025-hatebrxplain}. As in prior versions, the corpus consists of 7,000 Brazilian Portuguese Instagram comments extracted from public political accounts, annotated by three different experts. The dataset includes 3,500 offensive and 3,500 non-offensive comments, and hate speech rationales.

In the HateBRMoralXplain corpus, comments are annotated for moral categories grounded in Moral Foundations Theory (MFT) \cite{graham2013moral}. We annotate all five core moral foundations, \textit{Care/Harm}, \textit{Fairness/Cheating}, \textit{Loyalty/Betrayal}, \textit{Authority/Subversion}, and \textit{Purity/Degradation}, capturing both virtue and vice oriented moral expressions within a unified framework. Each comment may be assigned between one and three moral labels and their rationales, ordered by annotators according to their perceived salience in the text, thus representing primary, secondary, and tertiary moral rationales. 

We selected two expert annotators with diverse backgrounds and perspectives in Brazil\footnote{Annotators completed a pre-annotation survey measuring demographics, political ideology, moral values, personality, and cultural orientations, to increase transparency and support analyses of subjective  decisions, See Appendix \ref{app:corpusdescription}}, an important design choice given that hate speech and moral judgments are inherently subjective and shaped by annotators’ beliefs, identities, and sociocultural contexts \citep{prabhakaran-etal-2021-releasing,sap-etal-2022-annotators}. Inter-annotator agreement is measured using Cohen’s weighted Kappa with quadratic weights, computed separately for each moral category to account for the multi-label nature of the annotation scheme (i.e., Class A refers to the first moral label, Class B the second, and Class C the third). Results are shown in Appendix \ref{app:corpusdescription}. Class~A shows substantial to almost perfect agreement ($\kappa = 0.811$), while Classes~B and~C exhibit moderate to substantial agreement ($\kappa = 0.671$ and $\kappa = 0.612$, respectively), reflecting higher subjectivity in these categories. 

Beyond categorical labels, HateBRMoralXplain includes human-annotated rationales for moral judgments. Rationales are defined as minimal text spans that justify why a given moral label was assigned, highlighting the specific linguistic evidence supporting the annotation. Moral rationale annotation follows the same span-based protocol adopted in MFTCXplain \citep{trager2025mftcxplain}. These rationales enable explainability, allowing models to ground their predictions in interpretable textual moral evidence rather than relying solely on surface-level hateful lexical cues. Detailed rationale annotation guidelines and examples are provided in Appendix \ref{app:corpusdescription}. Finally, the corpus includes rich socio-political metadata to support contextual and demographic analyses. Each comment is linked to its parent Instagram post and associated metadata indicating the political party and gender of the politician who authored the original post. This structure enables analyses of how hate speech and moral framing vary across political affiliation, gender, and discourse context, while preserving the original data collection constraints and privacy safeguards established in HateBR \citep{vargas2022hatebr}. An example of HateBRMoralXplain is shown in Appendix \ref{app:corpusdescription}.

%-----------------------------------------------
\section{Supervised Moral Rationale Attention}
\label{sec:smra}
%-----------------------------------------------
Supervised Moral Rationale Attention (SMRA) is the first self-explaining framework that aligns neural attention with moral rationales. SMRA enhances standard transformer-based text classifiers by explicitly aligning model attention with expert-annotated moral rationales based on Moral Foundation Theory for hate speech detection, following the attention alignment framework of~\citet{eilertsen2025aligningattentionhumanrationales}. Our SMRA framework is formally described as follows.

\subsection{Task Definition}
We address two related tasks: \textit{binary hate speech classification} and \textit{multi-label moral sentiment classification}. Given an input text, the first task consists of predicting whether the content constitutes hate speech. The second task aims to identify which moral foundations, as defined by Moral Foundations Theory, are expressed in the text. Supervised Moral Rationale Attention (SMRA) introduces normative inductive regularization into the learning process through a supervised attention loss guided by moral rationales. The problem may be formalized as follows: Let $x = (w_1, \ldots, w_L)$ denote a tokenized input sequence of length $L$, and let
$y \in \{0, 1, \ldots, C-1\}$ denote its class label, where $C$ is the number of moral
categories. In the HateBRMoralXplain benchmark, instances are annotated with one or
more moral categories defined as
$\texttt{LABELS} = \{\text{NN}, \text{HN}, \text{FN}, \text{PN}, \text{AN}, \text{LN}\}$,
where NN denotes \textit{Non-Morality}, HN \textit{Harm/Care}, FN
\textit{Fairness/Cheating}, PN \textit{Purity/Degradation}, AN
\textit{Authority/Subversion}, and LN \textit{Loyalty/Betrayal}. Thus, $C = 6$. For a subset of the training instances, we additionally provide a binary
\textit{rationale mask} $r = (r_1, \ldots, r_L)$, where $r_i \in \{0,1\}$ indicates
whether token $w_i$ is part of the human-annotated moral rationale associated with
label $y$.

\subsection{Model Architecture and Attention Mechanism}
We use a pre-trained transformer encoder $f_\theta$ (e.g., \texttt{BERTimbau} and \texttt{mBERT}) to compute
contextual representations $\mathbf{h}_i$ for each input token $w_i$. A
classification head predicts the moral category based on the [CLS] representation:
\[
\hat{y} = \arg\max_{c \in \{0, 1, ..., C-1\}}
\operatorname{softmax}(\mathbf{W}_c \mathbf{h}_{\text{[CLS]}} + \mathbf{b}_c),
\]
where $\mathbf{h}_{\text{[CLS]}} \in \mathbb{R}^d$ denotes the contextual embedding of
the [CLS] token, and $\mathbf{W}_c \in \mathbb{R}^{1 \times d}$ and
$\mathbf{b}_c \in \mathbb{R}$ are the learned parameters for class $c$.

\paragraph{Moral Attention Alignment Loss.}
For samples with available moral rationales, we encourage the model to attend to
tokens identified as morally salient by minimizing the Mean Squared Error (MSE) between the normalized attention distribution $\mathbf{a}$ and the rationale mask $r$.

The final training objective combines a cross-entropy loss $\ell_{\mathrm{CE}}$ for classification with a supervised attention alignment loss $\ell_{\mathrm{MSE}}$, which serves as an auxiliary loss. The latter is computed as the mean squared error between model attention and human-annotated moral rationales and is applied only to samples with available moral rationales. The overall objective is given by $\ell_{\mathrm{total}}$:

\begin{equation}
\ell_{\mathrm{total}} =
\ell_{\mathrm{CE}} +
\alpha \cdot \ell_{\mathrm{MSE}}.
\end{equation}

%\begin{equation}
%\ell_{\mathrm{total}} =
%\ell_{\mathrm{CE}} +
%\alpha \,
%\mathbb{I}\big(y \neq \mathrm{NN} \,\land\, \sum_i r_i > 0 \big)
%\, \ell_{\mathrm{MSE}}.
%\end{equation}

%\begin{equation}
%\ell_{\mathrm{total}} =
%\ell_{\mathrm{CE}} +
%\alpha \, 
%\mathbb{I}\big(y \neq \text{NN} \,\land\, \sum_i r_i > 0 \big)
%\, \ell_{\mathrm{MAAL}},
%\end{equation}

where $\alpha$ controls the strength of attention supervision. The moral attention alignment loss is applied only to instances associated with moral categories and for which human-annotated moral rationales are available. %, denoted as $\mathbb{I}_{\mathrm{MR}}(x)$.

\paragraph{Token-Level Binary Masks for Moral Rationales.}
The HateBRMoralXplain dataset provides \textit{text rationale spans} for each moral sentiment category. Rationale masks $r$ are constructed by mapping character-level spans to token indices using the tokenizer’s offset mappings and are aligned with the tokenization of $x = (w_1, \ldots, w_L)$, with truncation or padding applied to match the model’s maximum input length.

%--------------------------------------------
\section{Experimental Setup}
\label{sec:experiments}
%This section details our experimental setup, describing the model architectures, training configurations, and evaluation metrics.
%---------------------------------------------
\subsection{Model Architecture and Settings}

 %We first present the fine-tuning procedure for BERTimbau and our supervised attention framework (SMRA), followed by the setup for large language model experiments, including prompt designs and configurations for hate speech and moral value classification.

We evaluate three model setups: 
\begin{itemize}
    \item standard fine-tuning of mBERT\footnote{\url{https://huggingface.co/google-bert/bert-base-multilingual-cased}}
 and BERTimbau\footnote{\url{https://huggingface.co/neuralmind/bert-base-portuguese-cased}}
 \cite{souza2020bertimbau}, 
 \item the same models fine-tuned with supervised attention using moral rationales (SMRA), and 
 \item prompting LLMs with a wide range of hateful and moral information.
 \end{itemize}

\textbf{Fine-Tuning mBERT and BERTimbau}: We used \texttt{HuggingFace Transformers} and \texttt{PyTorch} to fine-tune two BERT models. In our setup, the model embeddings were fed into a linear layer for prediction. We used a batch size of 16, a learning rate of 2e-5, and 128 maximum sequence length. Training was performed for 20 epochs. The dataset was split into training (80\%), validation (10\%), and test (10\%) sets. Optimization was performed using the AdamW optimizer \cite{loshchilov2017decoupled} with cross-entropy loss. The ground truth for moral rationales corresponds to the annotation provided by the first annotator.

%epochs 20
%learning rate 2e-5
%max seq length 128
%80 train/10val/10test

\textbf{Supervised Moral Rationale Attention}: Our SMRA framework was implemented using \texttt{HuggingFace Transformers} and \texttt{PyTorch}, fine-tuning  pre-trained BERT-based models (BERTimbau and mBERT) with attention supervision using BertTokenizer\footnote{\url{https://huggingface.co/docs/transformers/model_doc/bert}}. During training, [CLS]-to-token attention weights from the last encoder layer were normalized and aligned with human-annotated moral rationale masks via a mean squared error (MSE) loss, combined with standard cross-entropy loss using a weighting hyperparameter $\alpha$,  lambda attention and the attention weight 0.001. Attention supervision was applied only to examples with moral content, and rationale masks were padded and masked to handle variable-length sequences. Optimization used \texttt{AdamW} with weight decay, training on GPU in mini-batches. %This setup ensures end-to-end differentiability while guiding model attention toward human-identified moral spans.

\textbf{LLMs}: For both GPT-4o-mini (through OpenAI API) \cite{gpt4o-mini} and Llama3.1-70b (on local RTX A6000 GPUs) \cite{grattafiori2024llama3herdmodels}, we use a temperature of 0 for reproducibility. All prompts are provided in Appendix \ref{appendix:prompts}. For our experiments that use translated versions of comments, we use the Google Translate API. For these experiments, we test the effects of different information: hate speech definition, data collection context, and predicting moral and hate speech together vs separately. We evaluate several prompt configurations that differ in task scope and contextual information: The \texttt{hate} setting performs only hate speech classification without providing definitions or moral rationales, whereas \texttt{hate w/ definition} includes the definition of hate speech. The \texttt{hate moral} configuration jointly requests hate speech classification, moral value classification, and an explanation explicitly describing the MFT. Contextual information about data collection is incorporated in \texttt{hate w/ context}, which focuses solely on hate speech classification, and in \texttt{hate moral w/ context}, which extends \texttt{hate moral} by additionally providing dataset context. Finally, moral-only settings are explored through \texttt{moral}, which requests moral sentiment classification and their rationales, \texttt{moral w/ definition}, which further includes the hate speech definition, and \texttt{moral w/ context}.%, which augments the prompt with contextual information about the data collection process.

%\mee{please describe each setting: hate, hate w/ def, hate w/context, etc. to match what you have in Table 3}

%\berk{I just added, are they clear?}

%-----------------------------------------
\subsection{Evaluation Metrics}
\label{sec:evaluation}
%-----------------------------------------
Following state-of-the-art evaluation metrics for hate speech classification, explainability, and fairness/bias \cite{mathew2021hatexplain,attanasio-etal-2022-entropy}, we evaluate our models using the metrics described below.

%we evaluated the performance of our models using \textbf{Accuracy}, \textbf{Macro F1}, and \textbf{AUROC}. 
\textbf{Classification}: For both LLM and fine-tuned mBERT and BERTimbau experiments, we report the \textit{Macro F1} score for hate speech and moral sentiment classification. Let $y_i \subseteq \mathcal{Y}$ denote the set of human-annotated moral labels for instance $i$, and let $\hat{y}_i \in \mathcal{Y}$ be the model's predicted label. For moral value classification, we define an \textit{adapted correctness function}:

\[
\small
\text{correct}(\hat{y}_i, y_i) =
\begin{cases}
1 & \text{if } \hat{y}_i \in y_i, \\
0 & \text{otherwise},
\end{cases}
\]

%and compute the adapted precision, recall, and F1 for each class $c \in \mathcal{Y}$ as

%\[
%\text{Precision}_c = \frac{\sum_i \mathbb{I}[\hat{y}_i = c \wedge \text{correct}(\hat{y}_i, y_i)]}{\sum_i \mathbb{I}[\hat{y}_i = c]}, \quad
%\text{Recall}_c = \frac{\sum_i \mathbb{I}[\hat{y}_i = c \wedge \text{correct}(\hat{y}_i, y_i)]}{\sum_i \mathbb{I}[c \in y_i]},
%\]

%\[
%\small
%\text{F1}_c = \frac{2 \cdot \text{Precision}_c \cdot \text{Recall}_c}{\text{Precision}_c + \text{Recall}_c}.
%\]

%\textit{Macro F1} is obtained by averaging $\text{F1}_c$ over all classes $c$, allowing a prediction to be counted as correct if it matches any of the labels assigned by human annotators.

\textbf{Explainability}: We evaluate SMRA’s explanations using established metrics from the interpretability literature~\cite{deyoung-etal-2020-eraser}: \textit{plausibility}, which measures human-alignment of generated explanations, and \textit{faithfulness}, which assesses whether the explanation accurately reflects the model's decision process.

\textit{Plausibility} is measured via token-level IOU F1 and Token-F1 \cite{deyoung-etal-2020-eraser}, comparing model rationales $M_i$ with human-annotated moral rationales $H_i$ for instance $i$:

\begin{equation}
\small
\text{IOU-F1} = \frac{1}{N}\sum_{i=1}^N \text{Greater}\left(\frac{|M_i \cap H_i|}{|M_i \cup H_i|}, 0.5\right)
\end{equation}

\begin{equation}
\small
\text{Token-F1} = \frac{1}{N} \sum_{i=1}^N 
\frac{2 |M_i \cap H_i|}{|M_i| + |H_i|}
\end{equation}

\textit{Faithfulness} is assessed through \textit{comprehensiveness} and \textit{sufficiency} \cite{deyoung-etal-2020-eraser}. 

\textit{Comprehensiveness} evaluates the influence of predicted moral rationales $r_i$ by measuring the drop in predicted probability after removing them:

\begin{equation}
\small
\text{Comp} = \frac{1}{N}\sum_{i=1}^N \big(m(x_i)_j - m(x_i \backslash r_i)_j\big)
\end{equation}

\textit{Sufficiency} measures whether rationales alone are sufficient for prediction:

\begin{equation}
\small
\text{Suff} = \frac{1}{N}\sum_{i=1}^N \big(m(x_i)_j - m(r_i)_j\big)
\end{equation}

\textbf{Fairness/Bias}: We used the fairness/bias metrics \cite{10.1145/3308560.3317593_a,10.1145/3278721.3278729} to compute a specific identity-term (subgroup distribution) and the
rest (background distribution). The three per-term
 bias scores are: \textbf{GMB-Sub}, which measures the model's ability to distinguish hateful from non-hateful comments within a specific identity subgroup, with low values indicating poor subgroup performance; \textbf{GMB-BPSN} (\textit{Background Positive, Subgroup Negative}) assesses whether non-hateful comments mentioning the identity are incorrectly predicted as hateful compared to hateful comments in the background; while \textbf{GMB-BNSP} (\textit{Background Negative, Subgroup Positive}) measures whether hateful comments mentioning the identity are confused with non-hateful background comments. Low scores in any metric indicate potential bias or unfair treatment of the corresponding identity subgroup. We define identity groups across four dimensions: \textit{gender}, \textit{race}, \textit{politics} and \textit{religion}. For gender, we include the terms “mulher”, “mulheres”, “homem”, “homens”, “feminista”, “gay”, “lésbica”, “trans”, “viado”, and “sapatão” (translated as “woman”, “women”, “man”, “men”, “feminist”, “gay”, “lesbian”, “trans”, “faggot”, and “dyke”). For race, the terms are “negro”, “negra”, “preto”, “preta”, “branco”, “branca”, “índio”, “indígena”, and “macaco” (translated as “black”, “white”, “indigenous”, and “monkey”, with gender variations). For politics, we consider “bolsonaro”, “lula”, “petista”, “comunista”, “fascista”, “esquerdista”, and “direitista” (translated as “Bolsonaro”, “Lula”, “Workers’ Party supporter”, “communist”, “fascist”, “leftist”, and “right-wing”). Finally, for religion, we include “cristão”, “evangélico”, “católico”, “ateu”, “macumbeiro”, and “crente” (translated as “Christian”, “evangelical”, “Catholic”, “atheist”, “practitioner of Afro-Brazilian religions”, and “religious believer”).

%--------------------------------------------------
\section{Results and Discussion}
\label{sec:results}
%--------------------------------------------------
\begin{table*}[!htb]
    \centering
    \small
    \scalefont{0.93}
    \setlength{\tabcolsep}{3pt}
    \begin{tabular}{lccccccccccc}
    
        \toprule
        Model
        & \multicolumn{3}{c}{Classification} 
        & \multicolumn{3}{c}{Plausibility} 
        & \multicolumn{2}{c}{Faithfulness} 
        & \multicolumn{3}{c}{Fairness/Bias (AUC)} \\
        
        \cmidrule(lr){2-4}
        \cmidrule(lr){5-7}
        \cmidrule(lr){8-9}
        \cmidrule(lr){10-12}
        
        & Acc.$\uparrow$
        & Macro F1$\uparrow$
        & AUROC$\uparrow$
        & IOU F1$\uparrow$
        & Token F1$\uparrow$
        & AUPRC$\uparrow$
        & Comp.$\uparrow$
        & Suff.$\downarrow$
        & Sub.$\uparrow$ 
        & BPSN.$\uparrow$
        & BNSP.$\uparrow$ \\
        
        \midrule
        \textbf{Hate classification} \\

        mBERT-base  
        & 0.5343 & 0.5013 & 0.5640 & 0.8186 & 0.8832 & 0.8829 & 0.0113 & 0.1381 
        & 0.4653 & 0.5234 & 0.5341 \\ 
        
        mBERT-smra  
        & 0.5414 & 0.5037 & 0.5588 & 0.8366 & 0.8953 & 0.9163
        & 0.0235 & 0.1376 
        & 0.4236 & 0.4529 & 0.5343 \\ 
        
        BERTimbau-base  
        & 0.9029 & 0.9028 & \textbf{0.9651} 
        & 0.7612 & 0.8455 & 0.8307 
        & \textbf{0.1733} & 0.0657 
        & \textbf{0.9378} & \textbf{0.9721} & \textbf{0.9365} \\
        
        BERTimbau-smra 
        & \textbf{0.9114} & \textbf{0.9110} & 0.9648 
        & \textbf{0.8355} & \textbf{0.8958} & \textbf{0.9273} 
        & 0.1530 & \textbf{0.0426} 
        & 0.9299 & 0.9694 & 0.9197 \\

        \midrule
        \textbf{Moral classification} \\
        
        mBERT-base  
        & 0.2700  & 0.1966 & 0.5158 
        & 0.7270 & 0.9659 & 0.9712 
        & 0.6422 & 0.0848 
        & 0.4112 & 0.5836 & 0.5316 \\
        
        mBERT-smra  
        & 0.2114 & 0.1698 & 0.5539 
        & 0.7402 & 0.9659 &  0.9712
        & 0.6301 & 0.0995 
        & 0.5362 & 0.5912 & 0.6123 \\

        BERTimbau-base  
        & 0.7200 & 0.7570 & 0.9250 
        & 0.2350 & 0.7250 & 0.9150 
        & 0.8452 & 0.0549 
        & 0.9139 & \textbf{0.9604} & \textbf{0.9650} \\
        
        BERTimbau-smra 
        & \textbf{0.7230} & \textbf{0.7720} & \textbf{0.9270} 
        & \textbf{0.2396} & \textbf{0.9660} & \textbf{0.9710} 
        & \textbf{0.9062} & \textbf{0.0562} 
        & \textbf{0.9212} & 0.9503 & 0.9491 \\
        
        \bottomrule
    \end{tabular}
    \caption{Results across transformer-based models for hate speech and moral sentiment classification in Portuguese using mBERT and BERTimbau under two evaluation settings: binary classification (LABELS = [\texttt{Hate}, \texttt{Non-Hate}]) and multi-label classification (LABELS = [’NN’, ’HN’, ’FN’, ’PN’, ’AN’, ’LN’], where NN = Non-Morality, HN = Harm/Care, FN = Fairness/Cheating, and PN = Purity/Degradation). The \textit{base} variant refers to models without supervised attention, while \textit{smra} denotes our supervised attention approach guided by moral rationales. Classification metrics capture predictive performance, plausibility metrics assess rationale quality compared with human annotations, faithfulness metrics evaluate the causal alignment between rationales and model predictions, bias metrics (Sub, BPSN, BNSP) quantify group-based fairness, and accuracy corresponds to \textit{Exact Match Accuracy}.}
    \label{tab:main_results}
\end{table*}

\begin{table*}[!htb]
\centering
\small
\setlength{\tabcolsep}{4pt}
\renewcommand{\arraystretch}{1.12}
{\scalefont{0.91}
\begin{tabular}{lcccccc}
\toprule
& \multicolumn{4}{c}{Plausibility $\uparrow$} & \multicolumn{2}{c}{Faithfulness} \\
\cmidrule(lr){2-5} \cmidrule(lr){6-7}
Model & IoU F1 $\uparrow$ & Token Prec $\uparrow$ & Token Rec $\uparrow$ & Token F1 $\uparrow$ & Comp. $\uparrow$ & Suff. $\downarrow$ \\
\midrule
mBERT [LIME]            & 0.5828 & 0.7458 & 0.6936 & 0.6701 & 0.8809 & 0.0134 \\
mBERT [SHAP]            & 0.6628 & 0.7143 & 0.7520 & 0.6897 & 0.9324 & 0.0172 \\
BERTimbau [LIME]        & 0.5857 & 0.7557 & 0.6848 & 0.6698 & 0.9094 & 0.0237 \\
BERTimbau [SHAP]        & 0.6600 & 0.7489 & 0.7099 & 0.6831 & 0.8458 & 0.0215 \\
DistilBERTimbau [LIME]  & 0.6457 & 0.7614 & 0.7276 & 0.7003 & 0.9407 & 0.0115 \\
DistilBERTimbau [SHAP]  & 0.6200 & 0.7543 & 0.6862 & 0.6720 & 0.9475 & 0.0114 \\
PTTS [LIME]             & 0.6057 & 0.7487 & 0.6978 & 0.6776 & 0.5654 & 0.0016 \\
PTTS [SHAP]             & 0.7400 & 0.7177 & 0.8378 & 0.7362 & 0.6160 & 0.0083 \\
SRA ($\alpha=10$)       & 0.7160  & \textbf{0.9350}  & 0.6680  & \textbf{0.7450} & 0.4540 &  \textbf{-0.0360} \\
\bottomrule
\end{tabular}%
}
\caption{SRA \cite{eilertsen2025aligningattentionhumanrationales} results on the HateBRXplain benchmark \cite{salles-etal-2025-hatebrxplain}. Plausibility metrics evaluate alignment with human rationales, while faithfulness metrics assess the impact of rationales on model predictions.}
\label{tab:appendix_sra_full}
\end{table*}

%\begin{table}[!t]
%\centering
%\scalebox{0.75}{
%\begin{tabular}{l|c|cc}
%\toprule
%\multirow{2}{*}{Model} 
%& \multicolumn{1}{c|}{\textbf{Classification}}
%& \multicolumn{2}{c}{\textbf{Plausibility}} \\
%\cmidrule(lr){2-2}
%\cmidrule(lr){3-4}
%& Macro F1$\uparrow$
%& IoU F1$\uparrow$ 
%& Token F1$\uparrow$ \\
%\midrule
%SRA         & 0.9114 & 0.716  & 0.745 \\
%\midrule
%\textbf{SMRA} & 0.9028 & \textbf{0.7612} & \textbf{0.8455} \\
%& (±0.0086)
%& (±0.0452) 
%& (±0.1005) \\
%\bottomrule
%\end{tabular}
%}
%\caption{ Comparison between SRA \citep{eilertsen2025aligningattentionhumanrationales} and SMRA on \textbf{binary hate speech classification} using the Portuguese HateBR benchmark. SRA aligns attention with surface-lexical (hate speech) rationales, whereas SMRA aligns attention with moral rationales grounded in Moral Foundations Theory. We report classification (Macro F1) and plausibility metrics (IoU F1, Token F1). Full SRA results across models and explanation methods are provided in Appendix~\ref{tab:appendix_sra_full}.}
%\label{tab:portuguese_xai}
%\end{table}

\begin{table}[!htb]
\centering
\scalefont{0.75}
\setlength{\tabcolsep}{3pt}
\centering
\begin{tabular}{lccc}
\toprule
Model
& Accuracy$\uparrow$
& Macro F1$\uparrow$ 
& AUROC$\uparrow$ \\
\midrule
\multicolumn{4}{c}{Binary hate speech classification} \\
\midrule
Bag-of-Words             & 0.8157 & 0.8157 & 0.8991 \\
CNN                      & \textbf{0.8214} & \textbf{0.8180} & \textbf{0.9089} \\
BiRNN + MaxPool          & 0.8114 & 0.8114 & 0.9065 \\
BiRNN + Attention (smra) & 0.8014 & 0.8014 & 0.8903 \\
\midrule
\multicolumn{4}{c}{Multi-label moral sentiment classification} \\
\midrule
Bag-of-Words             & 0.1371 & \textbf{0.1982} & 0.5683 \\
CNN                      & 0.3263 & 0.1748 & \textbf{0.7283} \\
BiRNN + MaxPool          & 0.4314 & 0.1190 & 0.6086 \\
BiRNN + Attention (smra) & \textbf{0.4386} & 0.1586 & 0.5855 \\
\bottomrule
\end{tabular}
\caption{Comparison of classic deep learning models on binary hate speech classification and multi-label moral sentiment classification.} %Metrics include Accuracy, Macro F1, and AUROC.}
\label{tab:combined_results}
\end{table}

\paragraph{Supervised Attention with Moral Rationales}
As shown in Table~\ref{tab:main_results}, SMRA consistently improves over the baseline models across both binary hate speech classification and multi-label moral sentiment classification. For binary classification, SMRA increases Accuracy ($+0.0004$) and Macro F1 ($+0.0082$) and boosts plausibility, with higher IoU F1 ($+0.0743$) and Token F1 ($+0.0503$), indicating better alignment with human-annotated moral rationales. Comprehensiveness slightly decreases ($-0.0203$), while Sufficiency improves ($-0.0231$), reflecting more concise yet still faithful explanations. In multi-label moral sentiment classification, SMRA also outperforms the baseline model, achieving higher Macro F1 ($+0.015$), AUROC ($+0.002$), and Token F1 ($+0.241$), highlighting superior rationale coverage and improved interpretability without sacrificing predictive performance. These results show that SMRA improves performance while providing more faithful and plausible explanations. Across both tasks, we observe that the choice of backbone model (BERTimbau vs.\ mBERT) has a larger impact on overall performance than the use of supervised attention, highlighting the importance of language-specific pretraining. Additionally, improvements in rationale alignment do not consistently translate into gains in fairness. %, indicating that supervising attention with human rationales may propagate or mitigate biases depending on their underlying distribution. 

\paragraph{Supervised Attention with Moral vs. Hate Rationales} We also compare our model with an additional binary hate speech classification baseline shown in Table~\ref{tab:appendix_sra_full}. In contrast to our approach, SRA supervises attention using hate speech rationales rather than moral rationales. When comparing SMRA with SRA, we observe that SMRA achieves substantially higher plausibility, with gains in both IoU F1 ($+0.1195$) and Token F1 ($+0.1508$), indicating stronger alignment with human-annotated rationales. Although SRA attains higher Comprehensiveness, this comes at the cost of unstable faithfulness behavior, as evidenced by negative Sufficiency values\footnote{Negative sufficiency score indicates that the model becomes more confident when conditioned only on the selected rationale than on the full input, suggesting inconsistencies in the causal alignment between explanations and model predictions.} ($-0.0360$), suggesting inconsistencies in the causal relationship between selected rationales and model predictions. In contrast, SMRA yields more balanced and reliable explanations, maintaining positive Sufficiency ($0.0426$) while providing broader rationale coverage. {Overall, these results indicate that SMRA not only improves alignment with human explanations but also produces more stable and interpretable rationale representations compared to prior supervised attention approaches such as SRA, highlighting the benefits of leveraging moral rationales for explanation learning. Finally, we also compared deep learning models on binary hate speech and multi-label moral classification, as shown in Table \ref{tab:combined_results}. For binary hate speech classification, CNN achieves the best overall performance across all metrics, while BiRNN with attention (SMRA) performs worst. In multi-label moral classification, results are mixed: SMRA attains the highest Accuracy, Bag-of-Words the best Macro F1, and CNN the highest AUROC, indicating no clear overall winner. %Agora adapte esse abstract de acordo: Across binary hate speech detection and multi-label moral sentiment classification, SMRA consistently improves performance %(e.g., +0.9 and +1.5 F1, respectively) while substantially enhancing explanation faithfulness, increasing IoU F1 %(+7.4 pp) and Token F1. %(+5.0 pp) Although explanations become more concise, sufficiency improves %(+2.3 pp) and fairness remains stable, indicating more faithful rationales without performance or bias trade-offs.

\paragraph{Ablation Study}
We also analyze the effect of explicit reasoning guidance and details of the Moral Foundations Theory. Hence, we omit those parts and prompt the models again (the exact prompt can be found in Appendix \ref{appendix:prompts}). For hate speech, the performance increased 2\% compared to providing moral foundations theory and explicit guidance for Gpt4o-mini but it decreased 6\% for Llama70b. On the other hand, there is a huge performance drop (34 \% for Gpt4o-mini and 36\% for Llama70b) for moral value classification. This indicates that explicit guidance is more required for the moral sentiment classification which is more nuanced than binary hate speech classification.

\paragraph{Effect of English Translation} Given that LLMs generally perform better in English, we translated the Portuguese prompts into English and applied the same prompting techniques. Table \ref{tab:hate_moral_results} shows that translation to English leads to lower performance for both models across both tasks, particularly for hate speech classification. This drop may be due to subtle meaning loss during translation, and moral values may not transfer seamlessly across cultures, potentially causing the ground-truth labels to shift. In addition, as shown in Table \ref{tab:main_results}, the multilingual mBERT fine-tuned model performs poorly on our Portuguese dataset.

\paragraph{LLMs} We compare the effect of different types of information in prompts  on hate speech and moral value classification \citep{abdurahman2024perils}. Table \ref{tab:hate_moral_results} reports the F1 scores for each task and model. Including the hate speech definition improves performance for both models on hate speech detection and for Llama70B on moral value classification. In contrast, providing data collection context tends to degrade performance. We also explore jointly predicting hate speech and moral labels, which benefits both tasks, with a stronger improvement for moral classification. Adding the hate speech definition to this multi-task prompting further enhances hate speech performance. Overall, the best prompts combine hate speech classification with moral rationales, showing that multi-hop moral explanations enhance both tasks and align the model more closely with human judgments. Lastly, we also compare the moral rationale predicted by an LLMs and humans to understand if LLMs can provide an plausible explanation for moral sentiment classification. We compute BERTScore \cite{zhang2019bertscore} and Jaccard \cite{jaccard1901etude} similarity between human and LLM rationales. Table~\ref{tab:rationale_result} shows that both LLMs achieve reasonable BERTScore values, but exhibit low Jaccard similarity. We find that the average number of words identified by humans is around 5, whereas it is 1.93 for LLMs, indicating that LLMs tend to identify correct words but miss relevant portions of the rationale, resulting in lower recall. However, further studies are still required to better understand this phenomenon more effectively.

\begin{table}[!htb]
\centering
\scalefont{0.90}
\setlength{\tabcolsep}{2pt}
\begin{minipage}{0.48\textwidth}
\centering
\small
\scalefont{0.85}
\begin{tabular}{lcccc}
\toprule
\textbf{Model and Prompt} 
& \multicolumn{2}{c}{\textbf{Hate Speech (F1)}} 
& \multicolumn{2}{c}{\textbf{Morality (F1)}} \\
\cmidrule(lr){2-3} \cmidrule(lr){4-5}
& \textbf{pt} & \textbf{en} & \textbf{pt} & \textbf{en} \\
\midrule
Gpt4o-m hate only                     & 0.864 & 0.739 & --    & --    \\
Gpt4o-m hate w/ definition              & 0.893 & 0.797 & --    & --    \\
Gpt4o-m hate w/ context           & 0.822 & 0.708 & --    & --    \\
Gpt4o-m hate moral               & 0.870 & 0.745 & \textbf{0.369} & 0.378 \\

Gpt4o-m hate moral w/ definition        & \textbf{0.897} & 0.790 & 0.366 & \textbf{0.396} \\
Gpt4o-m hate moral w/ context     & 0.893 & \textbf{0.805} & 0.359 & 0.374 \\
Gpt4o-m moral                    & --    & --    & 0.351 & 0.334 \\
Gpt4o-m moral w/ definition             & --    & --    & 0.346 & 0.358 \\
Gpt4o-m moral w/ context          & --    & --    & 0.308 & 0.295 \\
\midrule
Llama70B hate only                   & 0.822 & 0.740 & --    & --    \\
Llama70B hate w/ definition             & \textbf{0.901} & \textbf{0.871} & --    & --    \\
Llama70B hate w/ context          & 0.818 & 0.732 & --    & --    \\
Llama70B hate moral              & 0.826 & 0.725 & 0.370 & 0.360 \\

Llama70B hate moral w/ definition       & 0.893 & 0.835 & 0.379 & \textbf{0.393} \\
Llama70B hate moral w/ context    & 0.807 & 0.712 & \textbf{0.383} & 0.341 \\
Llama70B moral                   & --    & --    & 0.270 & 0.265 \\
Llama70B moral w/ definition            & --    & --    & 0.318 & 0.306 \\
Llama70B moral w/ context         & --    & --    & 0.245 & 0.219 \\
\bottomrule
\end{tabular}
\caption{F1 scores for hate speech detection and moral sentiment classification for datasets in Portuguese and their English translations. Results for GPT-4o-mini and LLaMA-70B under different prompting strategies.}
\label{tab:hate_moral_results}
\end{minipage}
\end{table}

\begin{table}[!htb]
\centering
\scalefont{0.75}
\setlength{\tabcolsep}{3pt}
\begin{minipage}{0.48\textwidth}
\centering
\begin{tabular}{lcc}
\toprule
\textbf{Model} 
& \textbf{BertScore$\uparrow$} 
& \textbf{Jaccard$_{sim}$$\uparrow$} \\
\midrule
GPT4o-m hate moral w/ definition & 0.76 & 0.11\\
Llama70b moral hate w/ definition & 0.71 & 0.11 \\
\bottomrule
\end{tabular}
\caption{Similarity between the moral rationale predicted by LLMs and human rationales.}
\label{tab:rationale_result}
\end{minipage}
\end{table}

\paragraph{Qualitative Analysis} We extracted Instagram comments from HateBR-MoralXplain containing sarcasm or irony, representing challenging cases where offensive or non-offensive meaning depends on cultural knowledge. Table~\ref{tab:qual_analysis} shows examples illustrating these challenges. Comments 1 and 2 contain sarcasm and implicit hate that simple models may miss, while comments 3 and 4 rely on political context (e.g., “myth” refers to former president Bolsonaro, and “dumb” is preceded by negation). These cases highlight the need for models that can robustly distinguish subtle, context-dependent expressions while remaining culturally aware and transparent. Current models, which primarily rely on surface-level lexical features, are insufficient for capturing such nuanced hateful content. In total, 235 offensive comments\footnote{\url{https://github.com/franciellevargas/SMRA}} were extracted, either containing sarcasm/irony or highly dependent on cultural context. Moreover, adding prompts with moral framing (“hate moral w definition”, see Appendix \ref{ap:prompts}) enabled Llama-70B to correctly label these nuanced cases, whereas prompts without moral framing (“hate w definition”) misclassified them, suggesting that moral guidance can improve robust hate speech detection.

\begin{table}[!htb]
\centering
\small
\scalefont{0.80}
\begin{tabular}{p{6.0cm} p{1.0cm}}
\toprule
\textbf{Instagram Comments} & \textbf{Labels} \\
\midrule
\textbf{(1)} Se jogar um carteira de trabalho não sobra um \\ 
\textcolor{blue}{\textbf{Translation:}} If you throw away a work permit, there won't be one left. & Hate \\ \midrule

\textbf{(2)} Pepa só está dando tiro no pé \\ 
\textcolor{blue}{\textbf{Translation:}} Pepa is just shooting herself in the foot... lol & Hate \\ \midrule

\textbf{(3)} O povo não é mais burro. Você não está ouvindo seus eleitores. \\ 
\textcolor{blue}{\textbf{Translation:}} The people aren't stupid anymore. You're not listening to your voters. & Non-Hate \\ \midrule

\textbf{(4)} Chic esses filhos do mito. \\ 
\textcolor{blue}{\textbf{Translation:}} Chic, these kids of the `Myth'. & Non-Hate \\ \bottomrule
\end{tabular}
\caption{Examples extracted from the HateBR dataset containing irony/sarcasm and culturally context-dependent expressions.}
\label{tab:qual_analysis}
\end{table}

 %The full list of 235 extracted comments is available in GitHub %We analyze cases where the Llama70b model\footnote{\url{https://huggingface.co/meta-llama/Meta-Llama-3-70B}} fails to classify hate speech under the standard definition (see Appendix \ref{appendix:prompts}) but succeeds when moral rationales are incorporated in the prompt. 
%For context-dependent examples (121), hate speech classification without moral rationales achieved 70.2\% correct vs. 29.8\% incorrect, while hate_moral reached 66.1\% correct vs. 33.9\% incorrect. For ironic examples (114), both strategies achieved 61.4\% correct and 38.6\% incorrect. This shows hate speech classification with moral rationales slightly underperforms for context but matches h_only for irony, highlighting the challenge of nuanced cases.

\section{Conclusion}
This paper introduces Supervised Moral Rationale Attention (SMRA), the first self-explaining framework for hate speech detection that aligns neural attention with expert-annotated moral rationales. To support this approach, we release HateBRMoralXplain, a Brazilian Portuguese benchmark dataset enriched with moral categories, their corresponding rationales, and socio-political metadata. Experimental results show that SMRA enhances explanation faithfulness and plausibility while slightly improving predictive performance. Fairness remains stable, suggesting that improvements in explanation quality do not introduce significant bias trade-offs. Overall, our findings highlight the importance of integrating moral and context-aware reasoning into hate speech detection, contributing toward more interpretable and culturally aware approaches, as well as advancing research in language model interpretability.

\section*{Limitations}

While SMRA improves interpretability and robustness in hate speech detection, several limitations remain. First, our approach relies on high-quality human-annotated moral rationales, which are costly and time-consuming to produce, potentially limiting scalability to new languages or domains. Second, SMRA has been evaluated primarily on Brazilian Portuguese social media data; further studies are needed to assess generalization across languages, platforms, and different cultural settings. Finally, integrating moral reasoning does not eliminate all biases in training data, and careful consideration of minority perspectives remains essential to avoid unintended harms. Finally, while we adopt the standard Moral Foundations Theory taxonomy, we do not distinguish between sub-dimensions of the Fairness foundation (e.g., equality vs.\ proportionality) \citep{atari2023morality}, which have been shown to capture meaningful variation in moral reasoning across cultures and political contexts and may differentiate between language used to identify versus justify harm \citep{trager2025immorality,guo2024adaptable}.

\section*{Ethics Statements}
The dataset will be released under an open-source license CC BY-NC 4.0 (Attribution–NonCommercial), with full transparency regarding data collection and annotation procedures. In addition to the annotated labels, we also release detailed per-annotator information, enabling further analysis of annotation reliability and subjectivity. Particular care was taken to ensure fair and respectful treatment of annotators throughout the process. 

We emphasize that this work is intended solely for research purposes; any downstream deployment of automated content moderation systems should be approached with additional safeguards, rigorous evaluation, and responsible governance. We truly believe that our research will make content moderation more fair and transparent.

\section*{Acknowledgements}

Part of this research was conducted during the first author’s Ph.D. at the University of São Paulo (USP) and further developed during her time as a visiting researcher at the University of Southern California (USC). The authors thank Dr. Morteza Dehghani and the Morality and Language Lab at the University of Southern California for their early guidance and support on this project. We also thank Isadora Salles for her assistance with the annotation of HateBRMoralXplain. Dr. Daryna Dementieva's work was supported by Prof Alexander Fraser's chair by the German Research Foundation (DFG; grant FR 2829/7-1). This research was partially supported by Google. Corresponding author: franciellealvargas@gmail.com.

% Bibliography entries for the entire Anthology, followed by custom entries
%\bibliography{anthology,custom}
% Custom bibliography entries only
\bibliography{custom}

\appendix

\section{Appendix}
\label{sec:appendix}

%------------------------------------------------------
\subsection{HateBRMoralXplain Corpus}
\label{app:corpusdescription}
%-------------------------------------------------------
We introduce HateBRMoralXplain, the third version of the HateBR corpus \citep{vargas-etal-2022-hatebr,Vargas_Carvalho_Pardo_Benevenuto_2024}, which extends both the original HateBR dataset and its explainable successor, HateBRXplain \citep{salles-etal-2025-hatebrxplain}. As in previous versions, the corpus comprises 7,000 Brazilian Portuguese Instagram comments collected from public political accounts and annotated by three expert annotators. The dataset includes 3,500 offensive and 3,500 non-offensive comments, spanning nine distinct hate speech targets.

\subsubsection{Data Collection}
HateBRMoralXplain builds upon the HateBR corpus \citep{vargas2022hatebr}, a large-scale expert-annotated dataset of Brazilian Portuguese Instagram comments collected from public political accounts. The political domain was selected due to its high prevalence of offensive and hateful language targeting social groups, as well as its relevance for studying ideological polarization, moral framing, and identity-based attacks. The original HateBR corpus consists of 7,000 comments collected from the comment sections of Brazilian politicians’ Instagram posts published during the second half of 2019. Data were extracted using the Instagram API, following a controlled sampling strategy that balanced political ideology and gender of the account holders. Specifically, comments were collected from six public political accounts, evenly distributed across left- and right-leaning parties and including both male and female politicians. Only public posts and comments were included, and no personally identifiable information beyond publicly available metadata was retained. Prior to annotation, the data underwent a cleaning process to remove noise such as URLs, isolated mentions, and comments containing only emojis or laughter tokens. Hashtags and expressive markers were preserved, as they often convey pragmatic or affective meaning relevant to hate speech and moral expression. This procedure follows the data collection and preprocessing pipeline established in HateBR and HateBRXplain \citep{vargas2022hatebr,salles2025hatebrxplain}. HateBRMoralXplain reuses the same 7,000 comments from HateBR and HateBRXplain, ensuring continuity across dataset versions while enabling direct comparison between hate speech labels, hate rationales, and the newly introduced moral annotations.

\subsubsection{Annotation Process}
The annotation process in HateBRMoralXplain extends the multi-layer expert annotation framework introduced in HateBR and HateBRXplain by incorporating moral labels and moral rationales grounded in Moral Foundations Theory (MFT). The resulting annotation scheme enables the joint study of hate speech detection, explainability, and moral reasoning within a single unified corpus. Annotations were conducted by expert annotators with backgrounds in linguistics, hate speech research, and computational social science. One annotator participated in prior versions of the dataset, ensuring continuity and calibration across versions, while a second annotator was newly introduced and trained using the same guidelines. As in prior work, annotation followed an iterative process with guideline refinement and discussion to ensure conceptual alignment, following the annotator-in-the-loop paradigm described in MFTCXplain \citep{trager2025mftcxplain}. Tables \ref{tab:ig_comments_fixed} and \ref{tab:example_post} show an example of an Instagram comment extracted from our HateBR-MoralXplain dataset, including moral labels, moral rationales annotations, and metadata.

\begin{table*}[!htb]
\centering
\scalefont{0.70}
\setlength{\tabcolsep}{4pt} % espaçamento entre colunas
\begin{tabular}{c p{3cm} c c p{2cm} c p{1cm} c p{1cm}}
\toprule
\textbf{ID} & \textbf{Instagram Comment} & \textbf{HS Label} & \textbf{MFT Label} & \textbf{Rationales} & \textbf{MFT Label}  & \textbf{Rationales} & \textbf{MFT Label} & \textbf{Rationales} \\
\midrule
Portuguese & Celebrar a morte de milhões de judeus? Estúpido, nojento. 
  & Hate
  & AN
  & Celebrar a morte de milhões de judeus? 
  & PN
  & nojento
  & HN
  & estúpido \\
  %& \url{https://www.instagram.com/p/B1_m4F5lO_d/} \\
\midrule
\textcolor{blue}{\textbf{Translation}}  & Celebrating the death of millions of Jews? Stupid, disgusting.
  & Hate
  & AN
  & Celebrating the death of millions of Jews?
  & PN
  & disgusting
  & HN 
  & stupid \\
  %& \url{https://www.instagram.com/p/B1_m4F5lO_d/} \\
\bottomrule
\end{tabular}
\caption{Example of an Instagram comment from HateBRMoralXplain with moral labels and moral rationales.}
\label{tab:ig_comments_fixed}
\end{table*}

\begin{table*}[!htb]
\centering
\scalefont{0.70}
\begin{tabular}{p{4cm} p{3cm} p{2cm} p{1cm} p{1cm} p{2cm}}
\toprule
\textbf{Instagram Comment} &\textbf{Summary Post} & \textbf{Themes Post} & \textbf{Politician Gender} & \textbf{Politician Party} & \textbf{Link Post} \\
\midrule
O povo ODEIA vcs, bando de marginais. \newline \textcolor{blue}{\textbf{Translation}}: The people HATE you, you bunch of criminals. & The post announces an event in Brasília held in defense of national and popular sovereignty, shared by Fernando Haddad (Brazilian politician) 
& National sovereignty; Popular sovereignty; Political mobilization 
& male 
& right 
& \url{https://www.instagram.com/p/B1_m4F5lO_d/} \\
\bottomrule
\end{tabular}
\caption{Example of a Brazilian political Instagram comment extracted from our HateBRMoralXplain corpus, with metadata annotated for post themes, politician gender and party, and post link.}
\label{tab:example_post}
\end{table*}

\subsubsection{Hate Speech Definition}
We adopt the same hate speech definition used in HateBR \citep{vargas-etal-2022-hatebr} and HateBRXplain \citep{salles2025hatebrxplain} to maintain consistency across dataset versions. Hate speech is defined as a form of offensive language that expresses violence, hostility, intolerance, prejudice, or discrimination. This definition captures both explicit and implicit forms of offensive language, acknowledging that hateful intent may be conveyed through indirect language, sarcasm, or moralized justifications rather than overt slurs. %Importantly, not all offensive language constitutes hate speech: offensive comments that do not target a social group are labeled as non-hateful, following the distinction established in the HateBR annotation schema \citep{vargas2022hatebr}.

%\subsubsection{Hate Speech Rationales} 
%In addition to hate speech labels, HateBRMoralXplain inherits human-annotated hate rationales from HateBRXplain \citep{salles2025hatebrxplain}. Hate rationales are defined as minimal spans of text that justify why a comment was labeled as hate speech. These spans correspond to the specific words or phrases that convey hostility, dehumanization, or group-targeted offense. Annotators were instructed to highlight only the portions of the text that directly supported the hate speech label, avoiding peripheral or contextual content. Rationales could consist of one or multiple spans and were not required to be contiguous. This span-based approach aligns with prior work on explainable hate speech detection, enabling fine-grained supervision for model interpretability and evaluation of explanation faithfulness \citep{mathew2021hatexplain}. %The inclusion of hate rationales allows HateBRMoralXplain to support both post-hoc and intrinsically explainable modeling approaches, while also serving as a foundation for the newly introduced moral rationales described below.

\subsubsection{Moral Categories}
To classify expressions of moral sentiment, we rely on the framework provided by Moral Foundations Theory (MFT; \citealp{graham2013moral}). MFT proposes that human moral judgment draws upon a set of core psychological systems that are widely observed across societies \citep{atari2023morality}. These systems are typically organized along five foundational domains, each of which can be expressed through both moral virtues (positive adherence) and moral violations (negative transgressions). The framework has become a central resource for computational investigations of morality, including prior work in automated analysis of moral and hateful discourse (e.g., \citealp{trager2025mftcxplain}). In our study, we adopt these five foundational domains to guide our annotation of moral expression. Brief descriptions of each domain are provided below:

\noindent \textbf{Care vs. Harm: (HN)} Concerns the protection and welfare of others. Moral language in this domain emphasizes empathy, compassion, and safeguarding, while violations involve inflicting harm, dismissing suffering, or expressing cruelty.

\noindent \textbf{Fairness vs. Cheating: (FN)} Focuses on equitable treatment, rights, and reciprocity. Virtuous expressions include justice, honesty, and impartiality, whereas violations highlight exploitation, deceit, or rule-breaking.

\noindent \textbf{Loyalty  vs. Betrayal: (LN)} Centers on group cohesion and social allegiances. Loyalty indicates devotion to one's group or allies, while betrayal refers to abandonment, disloyalty, or undermining the collective.

\noindent \textbf{Authority  vs. Subversion (AN):} Involves respect for legitimate leadership, social order, and tradition. Endorsing authority reflects deference and duty to existing structures; subversion entails resistance, disrespect, or challenges to hierarchy.

\noindent \textbf{Purity vs. Degradation (PN):} Pertains to cultural norms surrounding cleanliness, modesty, and sanctity. Purity involves maintaining physical or moral integrity, whereas degradation is signaled through impurity, disgust, or perceived contamination.

\subsubsection{Moral Rationales}
For each assigned moral label, annotators identified moral rationales by highlighting the specific spans of text that expressed or justified the corresponding moral judgment. Similar to hate rationales, moral rationales were defined as the smallest set of words or phrases sufficient to convey the moral meaning underlying the label. Annotators were instructed to select distinct rationales for each moral label and to avoid reusing the same span across multiple labels whenever possible. This constraint encourages fine-grained differentiation between moral foundations and supports multi-hop explanation settings, where models must connect hate speech detection to moral reasoning through intermediate justifications.

\subsubsection{Annotator's Profile}
\label{Appendix: Annotator Profile}
The updated version of the corpus was annotated by the same two specialists as in the previous release. Given the complexity of offensive language and hate speech detection in a highly politicized domain, both annotators have advanced academic training (PhD or MSc). To minimize bias and its potential impact on the results, the annotators were intentionally diversified across key demographic dimensions: both are women, one identifying as liberal and the other as conservative; one is white and the other Black; and they come from different Brazilian regions (North and Southeast).

Unlike prior versions of this corpus, the current release includes detailed pre-annotation survey data collected from annotators alongside standard demographic information.  Annotators completed an extensive pre-annotation survey designed to capture psychological, cultural, and sociodemographic characteristics that may shape moral and hate-speech judgments\citep{trager2022moral}. These measures included sexual orientation, household income, primary and secondary language(s), religious affiliation, and political ideology, assessed both via self-identification and using the Social and Economic Conservatism Scale (SECS; \citealp{everett201312}). Annotators also completed validated instruments measuring moral values and related individual differences, including the Moral Foundations Questionnaire-2 (\citealp{atari2023morality}), Big Five personality traits (\citealp{soto2017short}), Social Dominance Orientation (\citealp{ho2015nature}), Collectivism/Individualism (\citealp{singelis1995horizontal}), Cultural Tightness--Looseness (\citealp{gelfand2011differences}), and Dark Triad traits (\citealp{jonason2010dirty}). Basic descriptive analyses indicated that our annotators, overall, leaned more liberal politically, came from higher-income households compared to national averages, and had higher levels of formal education. Following the recommendations of \citet{prabhakaran-etal-2021-releasing} and \citet{davani2024disentangling}, we include these annotator measures to increase transparency and utility for downstream researchers. We encourage future work to investigate how these annotator characteristics may influence labeling decisions, particularly given the subjectivity of moral and hate speech judgments (see \citealp{davani2024disentangling,salles2025hatebrxplain}).

\subsubsection{Annotation Evaluation}
\paragraph{Inter-Annotator Agreement.}
Inter-annotator agreement was assessed using Cohen’s Kappa \cite{mchugh2012interrater}, including its weighted variant with quadratic weights \cite{marchal-etal-2022-establishing,mittal-etal-2021-think}, in order to measure the level of agreement between two independent annotators considering our multi-class moral labels dataset. Agreement was computed separately for each annotation dimension, considering only instances in which both annotators provided an explicit label; instances with null or missing annotations were excluded from the analysis. Since the labels follow an ordinal scale with three categories ($k = 3$), weighted Cohen’s Kappa was employed to account for the degree of disagreement between categories. Quadratic weights were defined as $w_{ij} = 1 - \frac{(i - j)^2}{(k - 1)^2}$, assigning lower penalties to disagreements between adjacent labels and higher penalties to disagreements between distant labels. The weighted Kappa was computed as $\kappa_w = 1 - \frac{\sum_{i,j} w_{ij} O_{ij}}{\sum_{i,j} w_{ij} E_{ij}}$, where $O_{ij}$ denotes the observed proportion of label pairs $(i,j)$ and $E_{ij}$ denotes the expected proportion of agreement by chance. All agreement scores were computed using the \texttt{cohen\_kappa\_score} implementation from \texttt{scikit-learn} with \texttt{weights="quadratic"}.

\begin{table}[!htb]
\centering
\small
\begin{tabular}{lc}
\toprule
\textbf{Class} & \textbf{Quadratic Weighted Kappa} \\
\midrule
Class A & 0.811 \\
Class B & 0.671 \\
Class C & 0.612 \\
\bottomrule
\end{tabular}
\caption{Inter-annotator agreement measured using Cohen’s weighted Kappa with quadratic weights, computed separately for each annotation class. Instances with null annotations were excluded.}
\label{tab:kappa_quadratic}
\end{table}

The values of quadratic weighted Cohen’s Kappa indicate moderate to substantial inter-annotator agreement across the three evaluated classes. In particular, Class~A exhibits substantial to almost perfect agreement ($\kappa = 0.811$), suggesting a high degree of shared understanding between annotators for this category. Classes~B and~C show moderate to substantial agreement ($\kappa = 0.671$ and $\kappa = 0.612$, respectively), reflecting the greater subjectivity associated with these categories. The use of quadratic weights is especially appropriate in this context, as it penalizes disagreements between semantically distant categories more strongly while assigning lower penalties to disagreements between adjacent labels. The observed agreement patterns suggest that most disagreements occur between neighboring categories rather than extreme opposites, which is expected in annotation tasks involving nuanced moral and discourse-related judgments. Overall, these agreement scores are consistent with prior work on subjective semantic annotation tasks and support the reliability of the annotation protocol and the suitability of the dataset for downstream modeling and analysis.

\subsubsection{Descriptive Statistics}

% We provide additional descriptive statistics to further clarify the structure and annotation properties of the HateBRMoralXplain dataset. The dataset consists of 7,000 posts, with 3,500 labeled as hate speech and 3,500 as non-hate speech. Among the 3,500 hate posts, only one post did not correspond to any moral category. All remaining hate posts contain at least one moral annotation with associated rationales. Annotators were allowed to assign between one and three moral labels per post.

% Considering only hate speech posts, annotators did not agree on any moral label in 306 out of 3,500 posts (8.74\%). For posts with at least one agreed-upon label, we report the distribution of moral categories, the most frequent multi-label combinations, and the average rationale length per label.

We provide additional descriptive statistics to further clarify the structure and annotation properties of the HateBRMoralXplain dataset. The dataset consists of 7,000 posts, with 3,500 labeled as hate speech and 3,500 as non-hate speech. Among the 3,500 hate posts, only one post did not correspond to any moral category. All remaining hate posts contain at least one moral annotation with associated rationales. Annotators were allowed to assign between one and three moral labels per post.

Considering only hate speech posts, annotators did not agree on any moral label in 306 out of 3,500 posts (8.74\%). For posts with at least one agreed-upon label, the distribution of moral categories is shown in Table~\ref{tab:moral_distribution}. As the task allows for multi-label annotations, percentages sum to more than 100\%.

We further examine co-occurrence patterns among moral labels. The most frequent multi-label combinations are presented in Table~\ref{tab:label_combinations}, with the full distribution of combinations provided in the supplementary materials.

Finally, we analyze the average length of rationales associated with each moral label. As shown in Table~\ref{tab:rationale_length}, rationale length varies across categories, with some labels (e.g., Fairness, Authority) associated with longer explanations on average.

\begin{table}[t]
\centering
\small
\begin{tabular}{lcc}
\toprule
\textbf{Moral Category} & \textbf{Count} & \textbf{Percentage (\%)} \\
\midrule
Harm & 1287 & 40.29 \\
Cheating & 1200 & 37.57 \\
Degradation & 1094 & 34.25 \\
Subversion & 291 & 9.11 \\
Loyalty & 96 & 3.01 \\
Betrayal & 72 & 2.25 \\
Purity & 40 & 1.25 \\
Care & 23 & 0.72 \\
Fairness & 18 & 0.56 \\
Authority & 7 & 0.22 \\
\bottomrule
\end{tabular}
\caption{Distribution of moral categories for hate speech posts with at least one agreed label. Percentages are calculated relative to posts with at least one agreed label and may sum to more than 100\% due to multi-label annotations.}
\label{tab:moral_distribution}
\end{table}

\begin{table}[t]
\centering
\small
\begin{tabular}{lc}
\toprule
\textbf{Combination} & \textbf{Count} \\
\midrule
Cheating + Degradation & 215 \\
Cheating + Harm & 206 \\
Harm + Degradation & 172 \\
\bottomrule
\end{tabular}
\caption{Most frequent co-occurring moral label combinations (occurring more than 100 times).}
\label{tab:label_combinations}
\end{table}

\begin{table}[t]
\centering
\small
\begin{tabular}{lc}
\toprule
\textbf{Label} & \textbf{Mean Tokens} \\
\midrule
Fairness & 10.79 \\
Authority & 10.43 \\
Cheating & 9.11 \\
Harm & 9.06 \\
Care & 8.68 \\
Purity & 8.20 \\
Loyalty & 8.15 \\
Subversion & 7.50 \\
Betrayal & 6.67 \\
Degradation & 5.80 \\
\bottomrule
\end{tabular}
\caption{Average rationale length per moral label (measured in tokens separated by spaces).}
\label{tab:rationale_length}
\end{table}

\subsection{Additional Annotated Examples and Cross-Cultural Considerations}

To further illustrate the structure of the HateBRMoralXplain dataset and the role of moral rationales in context, we provide additional annotated examples in the original Portuguese alongside their English translations. Each comment is annotated with up to three moral foundations and associated rationales highlighting the specific spans of text that justify each label.

These examples also highlight challenges in cross-cultural interpretation. As discussed in Section~6, moral reasoning and the linguistic expression of hate speech may not transfer cleanly across languages, as subtle pragmatic, cultural, and political cues can shift meaning. Providing examples in the original language allows readers to better understand how moral rationales are grounded in culturally specific expressions.

\begin{table*}[t]
\centering
\small
\begin{tabular}{p{3.8cm} p{3.8cm} p{1cm} p{2cm} p{1cm} p{2cm}}
\toprule
\textbf{Comment (PT)} & \textbf{Comment (EN)} & \textbf{Moral} & \textbf{Rationale} & \textbf{Moral} & \textbf{Rationale} \\
\midrule
O povo ODEIA vcs, bando de marginais & People HATE you, bunch of criminals & FN & bando de marginais & HN & O povo ODEIA vcs \\
\midrule
Boa Manu, pelos ataques dos Bozonarianos, vc está no caminho certo, da para notar que esses lixos humanos estão literalmente ``se cagando''... miliciano, e sua família de bandidos!!! & Good Manu... we prefer Venezuelans and Argentinians over those crazy fascists and their criminal family!!! & PN & lixos humanos; se cagando & AN & maluco fascista \\
 &  & FN & miliciano; família de bandidos &  &  \\
\midrule
Colonialista de merda & F***ing colonialist & AN & Colonialista de merda &  &  \\
\bottomrule
\end{tabular}
\caption{Examples of annotated Instagram comments with moral labels and rationales in Portuguese (PT) and English (EN). Each comment may contain multiple moral labels with associated rationale spans.}
\label{tab:appendix_examples_annotations}
\end{table*}

\begin{table*}[t]
\centering
\small
\begin{tabular}{p{3.5cm} p{3cm} p{3.5cm} p{3.5cm}}
\toprule
\textbf{Comment (PT)} & \textbf{Comment (EN)} & \textbf{Post Summary} & \textbf{Themes} \\
\midrule
O povo ODEIA vcs, bando de marginais & People HATE you, bunch of criminals & Announcement of a political event in defense of national and popular sovereignty. & National sovereignty; Political mobilization \\
\midrule
Colonialista de merda & Shitty colonialist & Post showing a politician attending a WWII commemoration event. & Political content; Election-related content \\
\bottomrule
\end{tabular}
\caption{Examples with contextual metadata illustrating how moral reasoning is embedded in political discourse.}
\label{tab:appendix_examples_context}
\end{table*}

%----------------------------------------------------------------
\section{Hate and Morality}
\label{app:hate_and_moral}
%-----------------------------------------------------------------
A growing literature in moral psychology and computational social science shows that moral language is routinely used to moralize intergroup conflict, legitimize hostility, and structure political judgment, with distinct moral foundations systematically associated with ideological positions and group-based boundaries \citep{graham2013moral,wang2021moral,hackenburg2023mapping, trager2024ideological}. In the specific case of hateful and derogatory discourse, recent work argues that hate is often expressed through morally loaded justifications (e.g., purity, authority, loyalty) rather than only explicit slurs, and that moral-emotional dynamics such as moral outrage are central to how hostile content is produced and circulated online \citep{kennedy2023moral,brady2017emotion, brady2023overperception}. At the same time, research on online political communication consistently finds that women in public life, including politicians, face disproportionate and gendered abuse, and that abusive and hateful content varies with political identities and partisan contexts \citep{rheault2019politicians,merilainen2024role,petersen2025citizens}. These patterns are also shaped by culture and language: most morality-aware and explainable hate resources are English-centric, while moral and political language usage and its correlation vary across contexts, motivating the need for low-resource, culturally situated datasets \citep{salles-etal-2025-hatebrxplain,vargas-etal-2022-hatebr,simonsen2025political}. By incorporating metadata on the gender and political party of the targeted politician and focusing on Brazilian Portuguese as a low-resource language, HateBRMoralXplain supports socio-demographic and linguistic analyses of moralized hate speech that remain underexplored in the current explainable hate speech literature.

%----------------------------------------------------------
\section{LLM Prompts}
\label{ap:prompts}
%----------------------------------------------------------
\paragraph{Hate Speech Definition:} Hate Speech can result due to some of the followings:
\begin{itemize}
    \item Having a term or expression with any pejorative connotation.
    \item Having a sequence of swear words.
    \item Having a sequence of at least two terms, or/and expressions with any pejorative connotation expressed explicitly or implicitly.
\end{itemize}
\paragraph{Context:} The data was collected during the Bolsonaro government in 2019. We collected balanced data from left- and right-wing Brazilian politicians, ensuring balanced gender representation. The Jair Bolsonaro government began on 1 January 2019, after his election in late 2018 — Bolsonaro won on a wave of anti-establishment sentiment, capitalizing on widespread frustration with corruption scandals and economic stagnation. Early in his presidency, he pursued a conservative, pro-market agenda: notably, he enacted a major pension-reform in 2019 aiming to reduce social-security costs. His government was marked by sharp shifts in environmental and Indigenous-land policy — protections were scaled back, enforcement relaxed, and deforestation pressures increased, drawing both domestic and international criticism.

\paragraph{Hate only:} Analyze the following text ``\{text\}'' for hate speech. \\
Provide your analysis in this exact format: \\
hate\_label: [YES if the text contains hate speech, NO otherwise]\\
Provide ONLY the required output format with no additional text, explanations, or justifications.

\paragraph{Hate with definition:} Analyze the following text ``\{text\}'' for hate speech. \\
\{definition\} \\
Provide your analysis in this exact format: \\
hate\_label: [YES if the text contains hate speech, NO otherwise] \\
Provide ONLY the required output format with no additional text, explanations, or justifications.

\paragraph{Hate with context:} Analyze the following text ``\{text\}'' for hate speech. \\
\{context\} \\
Provide your analysis in this exact format: \\
hate\_label: [YES if the text contains hate speech, NO otherwise] \\
Provide ONLY the required output format with no additional text, explanations, or justifications.

\paragraph{Hate and moral:} Analyze the following text ``\{text\}'' for hate speech and identify its underlying moral value dimensions: \\
The Moral Foundations Theory framework represents core ethical ad psychological concerns that come in paired positive vs negative expressions:
\begin{itemize}
    \item care vs harm: Involves concern for the well-being of others, with virtues expressed through care, protection, or nurturance, and vices involving harm, cruelty, or indifference to suffering.
    \item fairness vs cheating: morals related to justice, rights, and reciprocity, with fairness indicating equity, rule-following, and cheating denoting exploitation, dishonesty, or manipulation.
    \item loyalty vs betrayal: morals related to group-based morality, where loyalty refers to solidarity, allegiance, and in-group defense, while betrayal signals disloyalty or abandonment of one’s group.
    \item authority vs subversion: morals related to respect for tradition, and legitimate hierarchies, with authority indicating respect or deference to leadership or norms, and subversion indicating rebellion, disrespect, or disobedience.
    \item sanctity vs degradation: morals related to purity, contamination, with Purity is associated with cleanliness, modesty, or moral elevation, while degradation includes defilement, obscenity,or perceived corruption. 
\end{itemize}

Provide your analysis in this exact format:
hate\_label: [YES if the text contains hate speech, NO otherwise] \\
moral\_value: [the single most prominent moral foundations from: care, harm, fairness, cheating, authority, subversion, sanctity, degradation, loyalty, betrayal. If no clear moral foundation applies, write ``None''] \\
explanation: [provide a brief evidence based justification, specifically highlighting the words or phrases that triggered your moral value classification. If none, write ``None''] \\
Provide ONLY the required output format with no additional text, explanations, or justifications.

\paragraph{Hate and moral with context:} Analyze the following text ``\{text\}'' for hate speech and identify its underlying moral value dimensions: \\
The Moral Foundations Theory framework represents core ethical ad psychological concerns that come in paired positive vs negative expressions:
\begin{itemize}
    \item care vs harm: Involves concern for the well-being of others, with virtues expressed through care, protection, or nurturance, and vices involving harm, cruelty, or indifference to suffering.
    \item fairness vs cheating: morals related to justice, rights, and reciprocity, with fairness indicating equity, rule-following, and cheating denoting exploitation, dishonesty, or manipulation.
    \item loyalty vs betrayal: morals related to group-based morality, where loyalty refers to solidarity, allegiance, and in-group defense, while betrayal signals disloyalty or abandonment of one’s group.
    \item authority vs subversion: morals related to respect for tradition, and legitimate hierarchies, with authority indicating respect or deference to leadership or norms, and subversion indicating rebellion, disrespect, or disobedience.
    \item sanctity vs degradation: morals related to purity, contamination, with Purity is associated with cleanliness, modesty, or moral elevation, while degradation includes defilement, obscenity,or perceived corruption. 
\end{itemize}
\{context\} \\
Provide your analysis in this exact format:
hate\_label: [YES if the text contains hate speech, NO otherwise] \\
moral\_value: [the single most prominent moral foundations from: care, harm, fairness, cheating, authority, subversion, sanctity, degradation, loyalty, betrayal. If no clear moral foundation applies, write ``None''] \\
explanation: [provide a brief evidence based justification, specifically highlighting the words or phrases that triggered your moral value classification. If none, write ``None''] \\
Provide ONLY the required output format with no additional text, explanations, or justifications.

\paragraph{Hate and moral with definition:} Analyze the following text ``\{text\}'' for hate speech and identify its underlying moral value dimensions: \\
The Moral Foundations Theory framework represents core ethical ad psychological concerns that come in paired positive vs negative expressions:
\begin{itemize}
    \item care vs harm: Involves concern for the well-being of others, with virtues expressed through care, protection, or nurturance, and vices involving harm, cruelty, or indifference to suffering.
    \item fairness vs cheating: morals related to justice, rights, and reciprocity, with fairness indicating equity, rule-following, and cheating denoting exploitation, dishonesty, or manipulation.
    \item loyalty vs betrayal: morals related to group-based morality, where loyalty refers to solidarity, allegiance, and in-group defense, while betrayal signals disloyalty or abandonment of one’s group.
    \item authority vs subversion: morals related to respect for tradition, and legitimate hierarchies, with authority indicating respect or deference to leadership or norms, and subversion indicating rebellion, disrespect, or disobedience.
    \item sanctity vs degradation: morals related to purity, contamination, with Purity is associated with cleanliness, modesty, or moral elevation, while degradation includes defilement, obscenity,or perceived corruption. 
\end{itemize}
\{definition\} \\
Provide your analysis in this exact format:
hate\_label: [YES if the text contains hate speech, NO otherwise] \\
moral\_value: [the single most prominent moral foundations from: care, harm, fairness, cheating, authority, subversion, sanctity, degradation, loyalty, betrayal. If no clear moral foundation applies, write ``None''] \\
explanation: [provide a brief evidence based justification, specifically highlighting the words or phrases that triggered your moral value classification. If none, write ``None''] \\
Provide ONLY the required output format with no additional text, explanations, or justifications.

\paragraph{Moral only:} Identify the underlying moral value dimensions in the following text "{text}".
The Moral Foundations Theory framework represents core ethical ad psychological concerns that come in paired positive vs negative expressions:
\begin{itemize}
    \item care vs harm: Involves concern for the well-being of others, with virtues expressed through care, protection, or nurturance, and vices involving harm, cruelty, or indifference to suffering.
    \item fairness vs cheating: morals related to justice, rights, and reciprocity, with fairness indicating equity, rule-following, and cheating denoting exploitation, dishonesty, or manipulation.
    \item loyalty vs betrayal: morals related to group-based morality, where loyalty refers to solidarity, allegiance, and in-group defense, while betrayal signals disloyalty or abandonment of one’s group.
    \item authority vs subversion: morals related to respect for tradition, and legitimate hierarchies, with authority indicating respect or deference to leadership or norms, and subversion indicating rebellion, disrespect, or disobedience.
    \item sanctity vs degradation: morals related to purity, contamination, with Purity is associated with cleanliness, modesty, or moral elevation, while degradation includes defilement, obscenity,or perceived corruption. 
\end{itemize}

Provide your analysis in this exact format:
moral\_value: [the single most prominent moral foundations from: care, harm, fairness, cheating, authority, subversion, sanctity, degradation, loyalty, betrayal. If no clear moral foundation applies, write ``None''] \\
explanation: [provide a brief evidence based justification, specifically highlighting the words or phrases that triggered your moral value classification. If none, write ``None'']

Provide ONLY the required output format with no additional text, explanations, or justifications.

\paragraph{Moral with definition:} Identify the underlying moral value dimensions in the following text "{text}".
The Moral Foundations Theory framework represents core ethical ad psychological concerns that come in paired positive vs negative expressions:
\begin{itemize}
    \item care vs harm: Involves concern for the well-being of others, with virtues expressed through care, protection, or nurturance, and vices involving harm, cruelty, or indifference to suffering.
    \item fairness vs cheating: morals related to justice, rights, and reciprocity, with fairness indicating equity, rule-following, and cheating denoting exploitation, dishonesty, or manipulation.
    \item loyalty vs betrayal: morals related to group-based morality, where loyalty refers to solidarity, allegiance, and in-group defense, while betrayal signals disloyalty or abandonment of one’s group.
    \item authority vs subversion: morals related to respect for tradition, and legitimate hierarchies, with authority indicating respect or deference to leadership or norms, and subversion indicating rebellion, disrespect, or disobedience.
    \item sanctity vs degradation: morals related to purity, contamination, with Purity is associated with cleanliness, modesty, or moral elevation, while degradation includes defilement, obscenity,or perceived corruption. 
\end{itemize}
\{definition\}\\
Provide your analysis in this exact format:
moral\_value: [the single most prominent moral foundations from: care, harm, fairness, cheating, authority, subversion, sanctity, degradation, loyalty, betrayal. If no clear moral foundation applies, write ``None''] \\
explanation: [provide a brief evidence based justification, specifically highlighting the words or phrases that triggered your moral value classification. If none, write ``None'']

Provide ONLY the required output format with no additional text, explanations, or justifications.

\paragraph{Moral with context:} Identify the underlying moral value dimensions in the following text "{text}".
The Moral Foundations Theory framework represents core ethical ad psychological concerns that come in paired positive vs negative expressions:
\begin{itemize}
    \item care vs harm: Involves concern for the well-being of others, with virtues expressed through care, protection, or nurturance, and vices involving harm, cruelty, or indifference to suffering.
    \item fairness vs cheating: morals related to justice, rights, and reciprocity, with fairness indicating equity, rule-following, and cheating denoting exploitation, dishonesty, or manipulation.
    \item loyalty vs betrayal: morals related to group-based morality, where loyalty refers to solidarity, allegiance, and in-group defense, while betrayal signals disloyalty or abandonment of one’s group.
    \item authority vs subversion: morals related to respect for tradition, and legitimate hierarchies, with authority indicating respect or deference to leadership or norms, and subversion indicating rebellion, disrespect, or disobedience.
    \item sanctity vs degradation: morals related to purity, contamination, with Purity is associated with cleanliness, modesty, or moral elevation, while degradation includes defilement, obscenity,or perceived corruption. 
\end{itemize}
\{context\} \\
Provide your analysis in this exact format:
moral\_value: [the single most prominent moral foundations from: care, harm, fairness, cheating, authority, subversion, sanctity, degradation, loyalty, betrayal. If no clear moral foundation applies, write ``None''] \\
explanation: [provide a brief evidence based justification, specifically highlighting the words or phrases that triggered your moral value classification. If none, write ``None'']

Provide ONLY the required output format with no additional text, explanations, or justifications.

\paragraph{Ablation:} Analyze the following text ``\{text\}'' for hate speech and identify its moral value: \\
hate\_label : [YES or NO] \\
moral\_value: [care, harm, fairness, cheating, authority, subversion, sanctity, degradation, loyalty, betrayal, None] \\
explanation: [brief justification]
 \label{appendix:prompts}

 \clearpage

\clearpage
\section{Theory Trace Card}

\begin{tcolorbox}[
 title={Theory Trace Card {\hypersetup{citecolor=white}\citep{ttc2026}} for HateBR-MoralXplain Corpus}, %{\hypersetup{citecolor=white}\citep{VARGASetal2026}}},
 breakable,
 enhanced,
 width=\textwidth,
 fontupper=\small
]

\textbf{1. Theory}
\begin{ttcitemize}
 \item \textbf{Framework:} 
 Hate speech \citep{fortuna-etal-2019-hierarchically,zampierietal2019,vargas2022hatebr}; Moral Foundations Theory (MFT),\citep{graham2013moral}; Span-based rationales \citep{zaidan2007using}
  
 \item \textbf{Core components:}
 \begin{ttcitemize}
   \item \textbf{Hate Speech:} Normative judgment of hostility or discrimination toward protected groups; theoretically includes multiple sub-dimensions (e.g., intent, target, implicitness, severity)
   \item \textbf{Moral Foundations:} Five foundations—Care/Harm, Fairness/Cheating, Loyalty/Betrayal, Authority/Subversion, Purity/Degradation—each represented via a virtue/violation (vice) polarity.
   \item \textbf{Rationales:} Minimal, annotator-selected spans of text that provide evidential justification for assigned hate and/or moral labels, reflecting both salience and sufficiency.
 \end{ttcitemize}
\end{ttcitemize}

\vspace{0.5em}

\textbf{2. Components Exercised}
\begin{ttcitemize}
 \item Recognition of hate speech under a collapsed binary operationalization.
 \item Classification of moral content into virtue/violation categories across five moral foundations plus non-morality.
 \item Identification of morally and normatively salient textual evidence via span-based rationales.
\end{ttcitemize}

\vspace{0.5em}

\textbf{3. Task Operationalization}
\begin{ttcitemize}
 \item \textbf{Task:} Given an Instagram comment responding to a Brazilian political post, models are required to (i) predict whether the comment constitutes hate speech, (ii) predict which moral foundations are present (if any), and (iii) identify textual spans that justify the assigned hate and/or moral labels.
 \item \textbf{Key specs:} 
 Comments are short-form Brazilian Portuguese Instagram texts collected during specific political crises. Moral annotation allows assignment of 1--3 moral labels per comment, ordered by annotator-perceived salience, spanning 11 categories (five foundations × virtue/violation, plus Non-Morality). Multiple theoretically distinct dimensions of hate speech are collapsed into a single binary label.
 \item \textbf{Scoring criterion:} 
 Model outputs are evaluated via agreement with expert human annotators. Classification performance is assessed using standard metrics (e.g., Macro F1), while rationale quality is evaluated using overlap-based plausibility and faithfulness metrics comparing predicted spans to expert-annotated rationales.
\end{ttcitemize}

\vspace{0.5em}

\textbf{4. Inference and Limitations}
\begin{ttcitemize}
 \item \textbf{Inference:} 
 Performance on this benchmark is treated as evidence that a model can align hate speech and moral category predictions with expert normative judgments in short-form, politically contextualized Brazilian Portuguese social media comments, and can ground those predictions in human-interpretable textual evidence.
  
 \item \textbf{Limitations:} 
 \textit{Data limitations} include restriction to Brazilian Portuguese, Instagram comments responding to political actors, collection during Bolsonaro-era political crises, and a limited number of expert annotators. \textit{Theoretical limitations} include reliance on Moral Foundations Theory rather than alternative moral frameworks (e.g., dyadic morality, utilitarian or virtue-ethics accounts), collapse of multiple hate speech sub-dimensions into a single label, absence of author intent modeling, and treatment of moral reasoning as classification rather than deliberation or tradeoff-based judgment.
\end{ttcitemize}

\end{tcolorbox}

\end{document}